\documentclass[twoside]{article}
\usepackage[accepted]{aistats2024}

\usepackage[utf8]{inputenc} 
\usepackage[T1]{fontenc}    
\usepackage{hyperref}       
\usepackage{url}            
\usepackage{booktabs}       
\usepackage{amsfonts}       
\usepackage{nicefrac}       
\usepackage{microtype}      
\usepackage{xcolor}         
\usepackage{amsmath}
\usepackage{amssymb}
\usepackage{bm}
\usepackage{enumitem}
\usepackage{subfig}
\usepackage{multirow} 
\usepackage{epsfig}
\usepackage{mathtools}
\usepackage{mismath}
\usepackage{algorithm}
\usepackage{algpseudocode}
\usepackage[round]{natbib}

\hypersetup{
    colorlinks,
    linkcolor={red!90!black},
    citecolor={blue!50!black},
    urlcolor={blue!80!black}
}

\makeatletter
\newcommand\RedeclareMathOperator{%
  \@ifstar{\def\rmo@s{m}\rmo@redeclare}{\def\rmo@s{o}\rmo@redeclare}%
}
\newcommand\rmo@redeclare[2]{%
  \begingroup \escapechar\m@ne\xdef\@gtempa{{\string#1}}\endgroup
  \expandafter\@ifundefined\@gtempa
     {\@latex@error{\noexpand#1undefined}\@ehc}%
     \relax
  \expandafter\rmo@declmathop\rmo@s{#1}{#2}}
\newcommand\rmo@declmathop[3]{%
  \DeclareRobustCommand{#2}{\qopname\newmcodes@#1{#3}}%
}
\@onlypreamble\RedeclareMathOperator
\makeatother

\DeclareMathOperator{\diag}{diag}
\DeclareMathOperator{\Tr}{Tr}

\RedeclareMathOperator{\E}{\mathbb{E}}
\DeclarePairedDelimiterX{\infdivx}[2]{[}{]}{%
  #1\;\delimsize\|\;#2%
}
\newcommand{\kldiv}{\text{KL}\infdivx}

\renewcommand{\geq}{\geqslant}

\renewcommand{\R}{\mathbb{R}}
\renewcommand{\N}{\mathcal{N}}
\renewcommand{\t}{\tau}

\newcommand\eqsp{\,}
\newcommand{\Lbb}{\mathbb{L}}
\newcommand{\ve}[1]{\mathbf{\bm{#1}}}  
\newcommand{\m}[1]{\mathbf{\bm{#1}}}  

\newcommand{\TP}{\mathcal{TP}}
\newcommand{\GP}{\mathcal{GP}}
\newcommand{\iup}[1]{{#1}^{(i)}} 
\newcommand{\gammab}{\boldsymbol{\gamma}}
\newcommand{\thetab}{\boldsymbol{\theta}}
\newcommand{\Sigmab}{\boldsymbol{\Sigma}}
\newcommand{\phib}{\boldsymbol{\phi}}

\newcommand{\lambdab}{\boldsymbol{\lambda}}
\newcommand{\Lambdab}{\boldsymbol{\Lambda}}

\newcommand{\Lcal}{\mathcal{L}}

\newcommand\rmd{\mathrm{d}}
\newcommand*{\tran}{^{\mkern-1.5mu\mathsf{T}}}

\newtheorem{Definition}{Definition}
\newtheorem{Proposition}{Proposition}
\newtheorem{Theorem}{Theorem}

\begin{document}

\twocolumn[

\aistatstitle{Identifiable Feature Learning for Spatial Data with Nonlinear ICA}

\aistatsauthor{Hermanni H{\"a}lv{\"a}$^{1}$ \And Jonathan So$^{2}$ \And  Richard E. Turner$^{2}$ \And Aapo Hyv{\"a}rinen$^{1}$}

\aistatsaddress{$^{1}$Department of Computer Science, University of Helsinki, Finland,\\ $^{2}$Department of Engineering, University of Cambridge, UK} ]

\begin{abstract}
Recently, nonlinear ICA has surfaced as a popular alternative to the many heuristic models used in deep representation learning and disentanglement. An advantage of nonlinear ICA is that a sophisticated identifiability theory has been developed; in particular, it has been proven that the original components can be recovered under sufficiently strong latent dependencies. Despite this general theory, practical nonlinear ICA algorithms have so far been mainly limited to data with one-dimensional latent dependencies, especially time-series data. In this paper, we introduce a new nonlinear ICA framework that employs $t$-process (TP) latent components which apply naturally to data with higher-dimensional dependency structures, such as spatial and spatio-temporal data. In particular, we develop a new learning and inference algorithm that extends variational inference methods to handle the combination of a deep neural network mixing function with the TP prior, and employs the method of inducing points for computational efficacy. On the theoretical side, we show that such TP independent components are identifiable under very general conditions. Further, Gaussian Process (GP) nonlinear ICA is established as a limit of the TP Nonlinear ICA model, and we prove that the identifiability of the latent components at this GP limit is more restricted. Namely, those components are identifiable if and only if they have distinctly different covariance kernels. Our algorithm and identifiability theorems are explored on simulated spatial data and real world spatio-temporal data. 
\end{abstract}

\section{INTRODUCTION}
Inferring semantically useful low-dimensional latent features from high dimensional real world data is a central goal in unsupervised feature learning. Many well-established approaches such as independent component analysis (ICA) and dictionary learning employ linear transformations, whereas recent approaches, such as nonlinear versions of ICA \citep{Hyva23pattern}, and variational autoencoders \citep{kingma2014autoencoding}, typically assume that the transformation from the latent to the observed space is highly nonlinear, often modeled as a deep neural network. 

Moving from linear to nonlinear generative models is likely to provide superior performance in many tasks, but also comes with challenges in interpretability and estimation. On the interpretability side, the problem is that many of these models are not \textit{identifiable} -- that is, they do not discover the ground-truth latent features, even in the limit of infinite data \citep{Hyva99NN,locatello19a}. Without access to the ground-truth latent features, we can expect worse performance and difficulties in interpreting results in several important tasks such as classification \citep{klindt2020towards,Banville21}, transfer learning \citep{Khemakhem20NIPS}, and causal inference \citep{Monti19UAI,wu2020causal}. In order to overcome these challenges, there has been an increased interest in developing identifiable generative models and the related theory. Much of this work has been in the context of nonlinear ICA as it would provide a way to perform \textit{principled} disentanglement learning, that is, without their usual problem of unidentifiability \citep{locatello19a,Hyva23pattern}. 

The central approach taken in such works has been to provide inductive biases that allow the models to be identified. The type of inductive biases used can be roughly divided into two groups. The first group assumes additional observed 'auxiliary' data \citep{Hyva16NIPS, Hyva19AISTATS, Khemakhem20iVAE}, such as audio that goes with the video data, or explicit knowledge about the time points at which non-stationary process switches distributions. Such auxiliary data is not always available, however. To cope with this, \citet{Hyva17AISTATS, Halva20UAI, halva2021disentangling} instead assume \textit{purely latent} structures that allow identifiability. These latent structures have been primarily different types of temporal dependencies, with applications on time-series data such as brain imaging. The only work that goes beyond temporal dependencies seems to be \citet{halva2021disentangling} who provide general theorems with \textit{sufficient} conditions that allow model identifiability with arbitrarily high dimensional latent dependencies, such as spatial data. Nevertheless, their work is mainly theoretical as the algorithm in \citet{halva2021disentangling} only applies to time-series data. In fact, no unsupervised learning method exists for performing non-linear ICA on data with more complex latent dependencies. This is a severe limitation as it hinders the applications of nonlinear ICA on many domains such as spatial data, where one would expect 2D dependencies, or spatio-temporal with 3D latent structures. These types of dependencies are ubiquitous, for example, in remote sensing and data used in meteorological, earth and other environmental sciences.

A well-known approach to modelling spatial dependencies would be to use Gaussian Processes (GP). However, the importance of non-Gaussianity is well-known in the linear ICA literature; whether this is the case in nonlinear ICA has not been thoroughly studied until now however. In this paper, we consider this via Student's t-Processes (TP)\citep{yuRobustMultitaskLearning2007, shah14_tp} of which GPs are a special case. In particular, we introduce Student's t-process nonlinear ICA (tp-NICA) which resolves the aforementioned limitations of temporal models by assuming TP latent components. Most importantly, unlike previous works, our model and the accompanying algorithm allow arbitrarily high dimensional latent dependencies. This makes the model particularly attractive for disentangling identifiable features from spatial data such as images or geographically structured data, as we will demonstrate in our experiments. 

Thus, the first major contribution of our paper is extending nonlinear ICA algorithms beyond time-series to spatial and other data with high-dimensional dependencies, without necessitating any additional auxiliary data. We also show the identifiability of the model by adapting the theory of \citet{halva2021disentangling}. Further, we establish GP nonlinear ICA (gp-NICA) as the limit of the tp-NICA model, but show that its identifiability is more constrained: the gp-NICA model is identifiable if and only if all the different processes have unique covariance kernels.

Our second major contribution is a new learning and inference algorithm for our tp-NICA model. In particular, our algorithm shows how to overcome two fundamental challenges: i) intractability that results from a non-conjugate non-exponential family latent prior combined with a nonlinear, neural network, observation likelihood, and ii) the expensive, cubic, computational cost typically associated with GPs and TPs. Our algorithm resolves these challenges by constructing a variational lower bound that utilizes the infinite GP-mixture representation of TPs, and by employing pseudo-points from sparse GP literature for computational efficacy, as well as other advances in variational inference.

\section{BACKGROUND} 
In this section we provide a brief technical background of some of the key concepts that we employ in our tp-NICA model and our learning and inference algorithm. 

\subsection{Identifiability and Nonlinear ICA}\label{back:identif}
A probabilistic model is considered identifiable if it is possible, at least in theory, to learn the model's ground-truth parameters. An unidentifiable model fails to satisfy this condition, and thus, any parameter estimates for such a model may not correspond to the ground-truth parameters. Formally, a model is identifiable up to some equivalence relation $\sim$ if
  $  p_{x}(\ve x; \ve \theta) = p_{x}(\ve x; \widetilde{\ve \theta}) \implies \ve \theta \sim \widetilde{\ve \theta},$
where $p_x(\ve x; \ve \theta)$ denotes the probability distribution of a random variable $\ve x$ parameterized by $\ve \theta$. Importantly for the present work, the most basic formulation of nonlinear ICA, namely a factorial prior on some latent components $p(\ve s) = \prod_{i=1}^N p(s^{(i)})$ along with a nonlinear mixing function $\ve x = \ve f(\ve s)$, has been shown to be catastrophically unidentifiable \citep{Hyva99NN} -- an infinite number of possible solutions exist. 
Several successful advances have been made over the recent years in developing alternative, identifiable, nonlinear ICA models. Of these, the most relevant to ours is \citet{halva2021disentangling}, whose Structured Nonlinear ICA (SNICA) framework defines a general set of sufficient identifiability conditions. These also apply to our model, and as such we will discuss them here in more detail. 
The SNICA framework utilizes a noisy version of the classic nonlinear ICA model \\
\begin{align}\label{eq:noisy_ica}
   \ve x_l = \ve f(\ve s_l) + \ve \varepsilon_l \eqsp \eqsp \eqsp \eqsp \forall_l \in \Lbb, 
\end{align}
where $\Lbb$ is an indexing set that could be arbitrarily high-dimensional (e.g. 2 dimensional for spatial data). \citet{halva2021disentangling} provide two main identifiability results. First, the noise-free mixture $\ve f(\ve s_l)$ can be identified from the noisy model, for arbitrary noise in \eqref{eq:noisy_ica}, under mild conditions that relate to the behaviour and non-Gaussianity of the noise-free mixture. Second, the ground-truth de-mixing function $\ve f^{-1}$ can be obtained from the distribution of $\ve f(\ve s_l)$ up to permutation and coordinate-wise transformations as long as conditions that govern sufficiently strong statistical dependency between index locations, and the condition of non quasi-Gaussianity, are satisfied. \citet{halva2021disentangling} also proposed a specific practical algorithm based upon their framework where the latent components follow autoregressive and hidden Markov models. While it was shown to be successful for temporal data, it cannot be applied on spatial data in any reasonable way, which is our main concern in this paper.

\subsection{Gaussian Processes and \texorpdfstring{$t$}{t}-Processes}
Gaussian process (GP) is a stochastic processes that specifies the joint distribution of infinitely many random variables such that any finite sample of those variables obeys a multivariate Gaussian distribution. Since GPs can represent the joint-distribution of infinitely many random variables, they are used as priors for continuous functions as well as latent processes. The dependencies between any variables in a GP are captured by a kernel function $\kappa(\ve x_i, \ve x_j)$, where $\ve x_i$ represents some input variable, such as location in space. For temporal data these could be one dimensional indices, while for spatial data they could, for example, be locations on a lattice. The crucial aspect, for this paper, is that the input variables can have an arbitrarily high dimension. GPs are also a limiting case of Student's-$t$ Processes (TP), as we show next.

If $\ve x \in \R^d$ follows a multivariate-$t$ distribution with  $\nu > 0$ degrees of freedom, we write ${\ve x\sim t_{\nu}(\ve \mu, \m \Sigma)}$, with its p.d.f given by
\begin{align}
    p(\ve x) = &\pi^{-\frac{d}{2}}\lvert \m \Sigma \rvert^{-\frac{1}{2}} \nu^{\frac{\nu}{2}} \frac{\Gamma(\frac{\nu+d}{2})}{\Gamma(\frac{\nu}{2})} \\ 
     & \times \left(\nu + (\ve x - \ve \mu)^T \m \Sigma^{-1} (\ve x - \ve \mu)\right)^{-\frac{\nu+d}{2}} \nonumber,
\end{align}
where $\Gamma(\cdot)$ is the Gamma function. When $\nu \rightarrow \infty$ this becomes a Gaussian p.d.f. Following \citet{yuRobustMultitaskLearning2007}, we can define $t-$Processes as follows:
\begin{Definition} (t-Process).
A random function $y: \R^d \rightarrow \R$ follows a t-Process ($\TP_{\nu}(h, \kappa)$) with $\nu > 0$ degrees of freedom, mean function $h(\cdot)$ and covariance kernel $\kappa(\cdot, \cdot)$, if for any $\ve x_1, \dots, \ve x_n \in \R^d$ with a finite $n > 0$:
\begin{align*}
    \ve y = (y(\ve x_1),\dots, y(\ve x_n))^T \sim t_{\nu}(\ve h, \m K),   
\end{align*}
where $\ve h = (h(\ve x_1), \dots, h(\ve x_n))^T$ and $\m K_{i,j}=\kappa(\ve x_i, \ve x_j)$.
\end{Definition}
It thus follows that a GP is the limit of a TP as the degrees of freedom parameter tends to infinity.

It is also well-known that a $t$-distribution can be re-framed as infinite mixture of Gaussian distributions with each Gaussian's covariance scaled by a Gamma distributed random variable. This extends also to $\TP$s as per the following proposition \citep{yuRobustMultitaskLearning2007}: 

\begin{Proposition} \label{prop:inf_mix} ($\TP$ as infinite mixture). $\TP_{\nu}(h, \kappa)$ can be sampled by repeatedly sampling a Gamma distributed random variable $\tau \sim \text{Gamma}(\frac{\nu}{2}, \frac{\nu}{2})$ followed by $\varphi \sim \GP(h, \frac{1}{\tau}\kappa).$
\end{Proposition}

 \subsection{Variational Inference} \label{sec:vi_background}
Our inference and learning algorithm combines several advances in variational inference and sparse GP literature and thus we provide a short background here. In the factored variational approach the posterior over a latent $\ve y$ is assumed to factorize over some partition $\mathcal{Z}_{1:M}$ such that $q(\ve y) := \prod_{j\in \mathcal{Z}_m} q_j(\ve y_j)$ is tractable, which allows a variational lower bound to be computed and optimized. Such approximations do not typically work for models with complex observation likelihoods such as neural networks. In black-box variational inference (BBVI) \citep{ranganath14} an approximate posterior distribution $q(\ve y_i \mid \ve \lambda_i)$ with free-form variational parameters $(\ve \lambda)_{i=1,\dots, m}$ is optimized. Each observation has its own free-form variational parameters, thus allowing for inference in complex distributions. The high variance of the gradients in BBVI is problematic however; \citet{rezende2014stochastic, kingma2014autoencoding} instead introduced the reparametrization trick which provides lower-variance gradients with respect to the variational parameters $\ve \lambda=(\ve \mu, \m \Sigma)$. Further, to deal with the intractable posterior $q(\ve y)$, VAEs \citep{kingma2014autoencoding} approximate it with a diagonal Gaussian: $\ve y_i \mid \ve x_i \sim \N(\ve m(\ve x_i), \ve \sigma(\ve x_i)^2 \m I)$ where $\ve m(\cdot), \ve \sigma(\cdot)$ are the outputs of an encoder neural network. 
Since VAEs assume factorial posteriors, they are not suited for models with complex latent dependencies. \citet{johnson2016composing} resolve this by assuming an approximate posterior in the form of $q(\ve y) \propto p(\ve y)l_{\ve \lambda}(\ve x \mid \ve y)$, where $l_{\ve \lambda}(\ve x \mid \ve y)$ is an approximate likelihood term chosen to be conjugate to the desired structured exponential family prior (such as a HMM). The approximate likelihood is again formed by an encoder neural network. This approach has inspired deep generative models with GP priors (see \citet{casale2018gaussian, pearce2020gaussian, ashman2020sparse, jazbec2021scalable}). For instance, \citet{ashman2020sparse} assume an approximate posterior:
\begin{align}\label{eq:sgp-vae}
    q(y^{(i)}) \propto p(y^{(i)})\prod_{t=1}^m \psi(\ve x_t^{(i)}; y^{(i)}, \Lbb)
\end{align}
with $q(\ve y) = \prod_i q(y^{(i)})$, where $y^{(i)} \sim \GP^{(i)}$, $(\ve x_m^{(i)})_{\forall m}$ are observations at $m$ locations of the indexing set $\Lbb$, where the approximate likelihood $\psi(\cdot)$ takes Gaussian form. Naive implementations of GPs suffer from cubic computational cost. Sparse GPs address this with the so called pseudo-points \citep{snelson_pseudo, quinonero-candela05a} which essentially summarize the GP at a much smaller number of pseudo-locations. These locations can be learned in a variational framework \citep{titsias09a, hensman2013gaussian, hensman2015scalable}. In practice many of the GP-VAE methods discussed above utilize these pseudo-points. For instance, \eqref{eq:sgp-vae} in \citet{ashman2020sparse} become $q(y^{(i)}) \propto p(y^{(i)})\prod_{t=1}^L \psi(\ve x_t^{(i)}; \ve u, \ve Z, \Lbb)$ where $\ve u$ are pseudo-observations at pseudo-locations $\ve Z$ and $L << m$.


\section{MODEL DEFINITION} \label{sec:model}
To define our proposed model, $t$-Process nonlinear ICA (tp-NICA), we start by assuming there exist $N$ statistically independent latent features $\ve s^{(i)} = (s_l^{(i)})_{l \in \Lbb}$ for $i \in \{1,\dots,N\}$. Here $\Lbb$  is an indexing set of arbitrary dimension. While previous theoretical works have sometimes allowed $\Lbb$ to also have any dimension, practical NICA algorithms have only allowed for one-dimensional indices, temporal index in particular. Here, our main application areas are spatial and spatio-temporal data and hence $\Lbb$ will usually be two- or three-dimensional, in other words, it's a subset of $\mathbb{N}^2$ or $\mathbb{N}^3$, so that we could equivalently write $\ve s^{(i)}=(s_{l}^{(i)})_{l\in \mathbb{N}^2} =(s_{k,l}^{(i)})_{k,l\in \mathbb{N}}$. Our algorithm will also be able to handle any other higher dimensional $\Lbb$ and thus is not limited to spatial or spatio-temporal data.

To allow for high-dimensional $\Lbb$, we assume that each of the independent components follows a $t$-Process (TP) over the indexing set $\Lbb$. TPs are a natural prior for independent components due to their non-Gaussianity and ability to model complex dependencies, but they have not previously been considered for NICA. Compared to GPs, they are also more flexible as the degree of freedom parameter controls the heaviness of the distribution's tails; allowing for heavy tails is useful as it is a characteristic that has been empirically observed in independent components \citep{hyvarinen2009natural}. Thus, we define 
\begin{align} \label{eq:tp_sample_nica}
    \ve s^{(i)} \sim \TP_{\nu^{(i)}}(h^{(i)}, \kappa^{(i)}),
\end{align}
with $h:\Lbb \rightarrow \R$ and $\kappa: \Lbb^2 \rightarrow \R$. The joint distribution of the independent components is defined as $p(\ve s_{l_1}, \dots, \ve s_{l_m})=\prod_{i=1}^N p(s_{l_1}^
{(i)}, \dots, s_{l_m}^{(i)})$ for $m \in \mathbb{N}^{\star}$ and $(l_1,\dots,l_m) \in \Lbb^m$. Additionally, we assume that the observed data is created from the latent components by a nonlinear mixing function that operates at each index of $\Lbb$ so that $\ve f : \R^N \rightarrow \R^M$ with $M \geq N$ is injective. We also assume observation noise denoted by $\ve \varepsilon_l \in \R^M$ that is i.i.d across the $M$ dimensions as well as with respect to the independent components. The noise can take any arbitrary distribution in theory but in practice we will usually assume a Gaussian distribution. As such, the observed data is generated as per Equation $\eqref{eq:noisy_ica}$ at each index $l \in \Lbb$. Note that this implies $p(\ve x_l | \ve s_l) = p_{\ve \varepsilon}(\ve x_l - \ve f(\ve s_l))$. Finally, the full joint distribution factorizes as
\begin{align} \label{eq:full_joint}
    p(\ve x_{1:m}, \ve s_{1:m})&=\prod_{j=1}^m p(\ve x_{l_j} \mid \ve s_{l_j})\prod_{i=1}^N p(\ve s^{(i)})
\end{align}
where we denote $s_{1:m} := s_{l_1},\dots, s_{l_m}$. 

Finally, if we take the degrees of freedom parameter $\nu$ in \eqref{eq:tp_sample_nica} to infinity, but otherwise keep the model as defined above, we establish the Gaussian Process NICA (gp-NICA) model in the limit and as a special case of the tp-NICA model 

\section{IDENTIFIABILITY} \label{sec:identif}
The tp-NICA model introduced above falls within the identifiability framework of \citet{halva2021disentangling}, thus leading to the following identifiability result: 

\begin{Theorem} (Identifiability of tp-NICA) \label{theo:main}
Assuming that the assumptions (A1), (A2) and (A3) of Theorem 1 in \citet{halva2021disentangling} apply, then the tp-NICA model is identifiable such that $p(\ve x; \ve f) = p(\ve x; \widetilde{\ve f}) \implies \ve f^{-1} \sim \widetilde{\ve f}^{-1}$, where $\sim$ denotes equivalence up to permutation and coordinate-wise bijective transformation of the elements of the de-mixing function $\ve f^{-1}$. 
\end{Theorem}

Our proof of this theorem (in Appendix \ref{apx:B}) relies on showing that $t$-processes are not (locally) quasi-Gaussian, which is a sufficient condition for identifiability \citep{halva2021disentangling}. 
This however does not preclude the possibility that GP components could lead to identifiable models in some instances. Indeed, \citet{halva2021disentangling} show that a noise-free variant of their model with GP independent components can be identified under the sufficient condition that the different GPs have distinct kernel functions. This raises the question whether non-Gaussianity is \textit{necessary} for identifiability. Our Theorem \ref{theo:nec} below shows that this is indeed the case. Subsequently, the gp-NICA model is identifiable if and \textit{only if} each component has a distinct covariance kernel, which is in contrast to tp-NICA where such necessary conditions do not apply. 

\begin{Theorem}\label{theo:nec} (Necessity of distinct covariance kernels with GP independent components). Assume that we have a model otherwise defined as in Section \ref{sec:model} except that in place of the $\TP$ distributed components, the model has $\GP$ distributed components: $\ve s^{(i)}\sim \GP(h^{(i)}, \kappa^{(i)}); i\in \{1,\dots,N\}$. Then $p(\ve x; \ve f) = p(\ve x; \widetilde{\ve f}) \implies \ve f^{-1} \sim \widetilde{\ve f}^{-1}$ if and only if the covariance kernels of the different components are unique so that $\kappa^{(i)}(s_l^{(i)}, s_m^{(i)}) \neq \kappa^{(j)}(s_l^{(j)}, s_m^{(j)}) \eqsp \eqsp \forall_{i \neq j}$.
\end{Theorem}
Note that Theorem \ref{theo:nec} considers the noise-free case because it is more general. To see this, notice that any necessary conditions for the noise-free case must be satisfied also by the noisy case. Our proof (in Appendix \ref{apx:B}) follows directly by extending a related theory for linear ICA by \citet{Belo97}.

\section{LEARNING AND INFERENCE}
The marginal log-likelihood of the tp-NICA model can be written as: 
\begin{align}\label{eq:ml}
	\Lcal \coloneqq \log p(\ve x) = \log \int_{\ve s}\prod_{j=1}^m p(\ve x_{l_j} \mid \ve s_{l_j})\prod_{i=1}^N p(\ve s^{(i)})  \eqsp \rmd \ve s,
\end{align}
Maximization of this marginal log-likelihood is clearly intractable:  $p(\ve x_l\mid \ve s_l)$ is a nonlinear observation likelihood parameterized by a deep neural network and Gaussian output noise, whilst $p(\ve s^{(i)})$ is a TP which is non-exponential family and thus non-conjugate, making it difficult to approximate $p(\ve x_l\mid \ve s_l)$ with a conjugate-likelihood (recall Section \ref{sec:vi_background}). Another challenge, as with GPs, is the apparent cubic computational cost. 

Therefore, we make two alterations to \eqref{eq:ml}. First, as a step towards tractability, recall Proposition \ref{prop:inf_mix}. This allows us to write the TP priors as infinite mixtures of GPs scaled by gamma random variables $\ve \tau$. Second, we introduce additional pseudo-variables, $\ve u$, to control the computation complexity of our algorithm. These pseudo-variables can be considered as additional latent variables at some extra input locations (Section \ref{sec:vi_background} and Appendix \ref{apx:distrib_latents}). Importantly, note that the $\ve u$ are \textit{not} observed and thus no additional data is needed; they are merely mathematically convenient additional latent variables that we integrate out. With these additions we can compute a tractable lower bound to our new marginal likelihood (see Appendix \ref{apx:A} for all derivations):
\begin{align}\label{eq:bounds}
	\Lcal &\coloneqq \log \int_{\ve{\t}} \int_{\ve s} \int_{\ve u}p(\ve x, \ve s, \ve u, \ve{\t}) \eqsp \rmd \ve u \rmd \ve s \rmd\ve \t \nonumber\\
 &\geq \E_{q(\ve \t)}\left[\E_{q(\ve s, \ve u \mid \ve \t)}\left[\log p(\ve x \mid\ve s, \ve u)\right]\right. \nonumber \\
 &\left.-\kldiv{q(\ve s, \ve u \mid \ve \t)}{p(\ve s, \ve u \mid \ve \t)}\right] - \kldiv{q(\ve\t)}{p(\ve{\t})} \nonumber \\
 &\geq \E_{q(\ve \t)}\underbrace{\left[\E_{\tilde{q}(\ve s \mid \ve \t)}\left[\log p(\ve x \mid \ve s)\right]\right.}_{(\ast)}\nonumber\\
 &-\underbrace{\left.\kldiv{q(\ve u \mid \ve \t)}{p(\ve u \mid \ve \t)}\right]}_{(\ast\ast)}] - \underbrace{\kldiv{q(\ve\t)}{p(\ve{\t})}}_{(\ast\ast\ast)},
\end{align}
where we denote $\tilde{q}(\ve s\mid \ve \t) \coloneqq \int p(\ve s \mid \ve u, \ve \t)q(\ve u \mid \ve \t) \rmd \ve u$. The second bound is due to our choosing to approximate $q(\ve s \mid \ve u, \ve \t)\coloneqq p(\ve s \mid \ve u, \ve \t)$, which allows for the computational benefits of using the pseudo-points. Further, we choose the following non-factorial posterior approximation for the pseudo-latents:
\begin{align}\label{main:approx}
	q(\ve u, \ve \t) =& q(\ve u\mid \ve \t; \phib,  \thetab)q(\ve \t; \phib)   \nonumber \\
	\coloneqq& \frac{\prod_{j=1}^J \psi(\ve u_j; \phib_j)\prod_{i=1}^N p(\iup{\ve u} \mid \iup{\t}; \iup{\thetab})}{Z(\phib,\thetab, \ve \t)}\\
    &\times\prod_{i=1}^N q(\iup{\t}; \iup{\phi}),\nonumber
\end{align}
where $J$ is the number of pseudo-locations, $\ve u_j = (u_{l_j}^{(i)})_{i=1:N}$, and $\ve u^{(i)} = (u_{l_j}^{(i)})_{j=1:J}$. Crucially, separating the TP prior into Gamma and $\GP$ distributions has now allowed us to approximate the previously intractable pseudo-likelihood term with a conjugate Gaussian factor $\psi(\ve u_j; \phib_j) = \exp\left\{-\frac{1}{2}(\ve u_j\tran \tilde{\ve W}_j\tran \tilde{\ve W}_j\ve u_j -2\ve u_j\tran   \tilde{\ve m}_j)\right\}$, where $(\tilde{\ve m}_j, \tilde{\ve W}_j) \in \phib_j$ are free-form variational parameters akin to \citet{ranganath14}, in particular $\tilde{\ve W}$ are parameterized as Cholesky factors. Notice that in \eqref{main:approx}  the approximate Gaussian factor and $p(\iup{\ve u} \mid \iup{\t}; \iup{\thetab})$ factorize over different indices, and consequently the resulting Gaussian $q(\ve u\mid \ve \t; \phib, \thetab)$ has a sparse, but non-factorized, form, which is in contrast to previous works that assume various limiting factorizations of the posterior. 

\begin{algorithm*}
\caption{Estimate tp-NICA lower bound and take gradients}\label{alg:main}
\begin{algorithmic}
\Require model parameters $\thetab:=(\thetab^{(i)})_{i=1,\dots,N}$, variational parameters $\phib:=((\iup{\phib})_{i=1,\dots,N}, \m Z)$, data $\ve x$, number of posterior $\ve \tau$ samples $N_t$, number of posterior $\ve s$ samples $N_s$
\Function{ELBOGradients}{$\thetab, \phib, \ve x, N_t, N_s$}
   \State $\{\widehat{\ve \tau}_t\}_{t=1:{N_t}} \underset{i.i.d}{\sim} q(\ve \tau; \phib)$ \Comment{$N_t$ Samples from approximate posterior}
   \State $\{\Lcal_t\}_{t=1:{N_t}} ,\eqsp \eqsp \{\widehat{\m S}_t\}_{t=1:{N_t}} \gets \Call{ConditionalInference}{\{\widehat{\ve \tau}_t\}_{t=1:{N_t}}, \thetab, \phib, \ve x, N_s}$ 
  \State $\Lcal \gets \frac{1}{N_t}\sum_{t=1}^{N_t}\Lcal_{t} - \kldiv{q(\ve\t; \phib)}{p(\ve{\t}, \gammab)}$ \Comment{Approximates \eqref{eq:bounds}}
  \State $\nabla_{\thetab}\Lcal, \nabla_{\phib}\Lcal \gets$ Autograd($\Lcal, \thetab, \phib$) \Comment{Compute gradients with re-parametrization trick}
  \State \Return $\Lcal, \nabla_{\thetab}\Lcal, \nabla_{\phib}\Lcal, \{\widehat{\m S}_t\}_{t=1:{N_t}}$
\EndFunction
\Function{ConditionalInference}{$\widehat{\ve \tau}_t, \thetab, \phib, \ve x, N_s$}
    \State $\{\widehat{\m s}_{s}\}_{s=1:N_s} \underset{i.i.d}{\sim} \tilde{q}(\ve s\mid \widehat{\ve \tau}_t; \thetab, \phib)$ \Comment{$N_s$ samples as per Sec. \ref{apx:compute_ast}}
    \State $\Lcal_t \gets \frac{1}{N_s}\sum_{s=1}^{N_s}\log p(\ve x \mid \widehat{\m s}_{s}) - \kldiv{q(\ve u \mid \widehat{\ve \tau}_t; \thetab)}{p(\ve u \mid \widehat{\ve \tau}_t; \phib)}$ \Comment{KL as per Sec \ref{apx:compute_astast}}
    \State \Return $\Lcal_t,\eqsp \eqsp \widehat{\m S}_t$  \Comment{$\widehat{\m S}_t:=\{\widehat{\m s}_{s}\}_{s=1:N_s}$}
\EndFunction
\end{algorithmic}
\end{algorithm*}

The final lower bound in \eqref{eq:bounds} allows sufficient tractability to estimate our model parameters and to perform inference. Terms $(\ast\ast), (\ast\ast\ast)$ can be computed in closed-form as they are just KL divergence between $\GP$ and Gamma distributions, respectively. The data-likelihood denoted $(\ast)$ is parameterized by a decoder network and thus remains intractable so that we take the VAE approach of reparametrized gradients after sampling from  $\tilde{q}(\ve s\mid \ve \t)$ which is a Gaussian. Finally, we also take reparameterized gradients of the outer expectation over the Gamma distributed $q(\iup{\t}; \iup{\phi})_{\forall_i}$. In practice, we use autograd to take gradients through all of \eqref{eq:bounds} and then optimize all the model parameters $\thetab$ and the free-form variational parameters $(\phib, \m Z)$ with stochastic gradient descent, where $\m Z$ are locations of the pseudo-latents. A step-by-step description is detailed in Algorithm \ref{alg:main}. For gp-NICA the Gamma expectation and KL-divergence terms in equation \eqref{eq:bounds} disappear, but otherwise the same algorithm can be utilized.

\section{EXPERIMENTS} \label{sec:xperiment}
In this section we explore the performance of the tp-NICA and gp-NICA models on simulated spatial as well as real world data. Our code will be openly available at [redacted for anonymity]. 

\begin{figure}
    \centering
    \vspace{.3in}
    \includegraphics[width=\columnwidth]{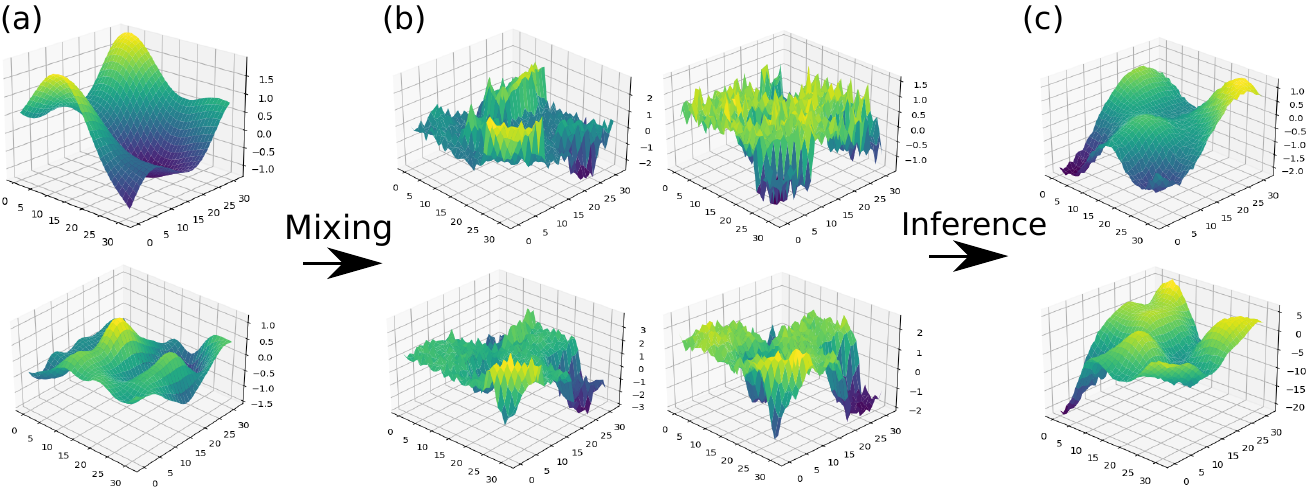}
    \caption{A simple illustration of (a) two TP independent components, (b) their four-dimensional mixture + observation noise (c) the inferred components at the end of tp-NICA training with MCC$=0.95$. Three mixing layers were used. Comparison of (a) and (c) illustrates well how the recovered components are identifiable up to component-wise bijective transformations.}
    \label{fig:example}
\end{figure}

\subsection{Experiments on Simulated Data}
We first tested how well the algorithms are able to estimate the ground-truth latent components from simulated spatial data. Since the identifiability theorems hold in the limit of infinite data and universal consistent estimators, it is of interest to see how these theorems are obeyed in practice in scenarios where the data and the estimator methods stray away from such theoretical ideals. For the simulation, we sampled spatial $32 \times 32$ latent components on a regular lattice, using a squared-exponential kernel. Both tp-NICA and gp-NICA were estimated on their own data. We considered specifications with 6 and 12 latent components, which were mixed into 12 and 24 32x32 observations, respectively, using a mixing neural network. The depth of the neural network was varied from one to four layers to simulate different levels of nonlinearity (one layer of mixing corresponds to linear ICA). Two data sets were generated, one where the squared exponential kernels' parameters were engineered to be distinctly different for each component, and another where the kernels for all the components were forced to be the same. Identifiability was measured by the mean absolute correlation coefficient (MCC) between the ground truth and estimated independent components. Figure \ref{fig:example} illustrates the type of data generated and displays an example of successful demixing by the tp-NICA model. 

None of the previous NICA algorithms are able to exploit spatial dependencies so there is no natural competitive benchmark. Nevertheless, we have included linear ICA and iVAE \citep{Khemakhem20iVAE} as naive baselines. These both models treat the spatial locations as independent. Further, iVAE requires some observed auxiliary data, which here we have taken to be the locations of the latent variables (i.e. the same input as into the GP kernels). 

\begin{figure}
    \centering
    \includegraphics[width=\columnwidth]{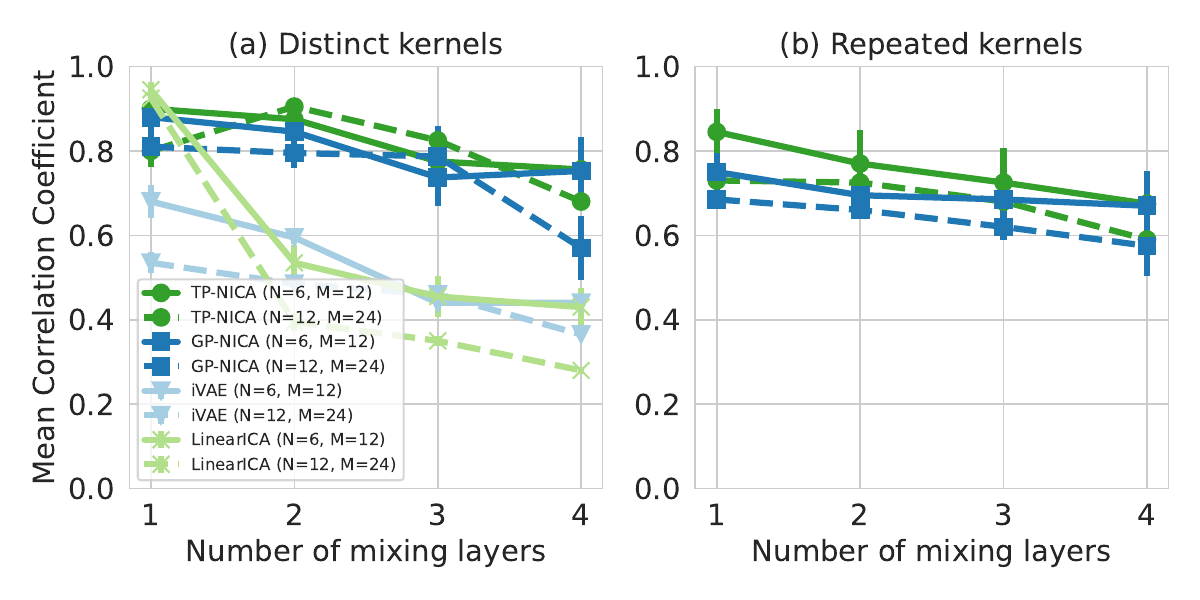}
    \caption{Mean absolute correlation coefficients between ground-truth independent components and their estimates for tp,gp-NICA, and iVAE and linear ICA baselines. gp-,tp-NICA are estimated on their own data, while the baseline models were estimated on data from both models with nearly identical results. 
    Results shown for varying levels of non-linearity (number of layers) for (a) distinctly different covariance kernels between the components, and (b) components with equivalent kernels.}
    \label{fig:sim}
\end{figure}
Our results in Figure \ref{fig:sim} largely correspond to theory: For data where kernels are forced to have distinctly different parameters, both tp-NICA and gp-NICA perform well, and much better than the naive baselines. The performance deteriorates slowly for all models as nonlinearity increases, which is to be expected due to limited data, computation, and estimation methods. For the setting with equal kernels across components, tp-NICA has 0.05-0.15-points higher MCC, showing that it's indeed more identifiable in this scenario as theory predicts. However, in contrast to theory, the gp-NICA model is not completely unidentifiable and as the number of mixing layers increases the two models performance converges. This is likely because in practice constraints such as the use of specific kernel and specific MLP mixing function imposes inductive biases which is in contrast to the theoretical assumptions of arbitrary mixing function and arbitrary covariance kernels. Also, in the most nonlinear setting, the difficulty of estimating the mixing likely dominates the results, rather than identifiability. Further implementation details and additional simulated experiments are found in the Appendix.

\subsection{Experiments on Real Spatio-temporal Data}
We also tested our models on real spatio-temporal data. In particular, we used data from the Computer Vision for Agriculture (CV4A) Kenya Crop Type dataset \citep{cv4a}. This dataset was originally designed for an entirely different supervised classification task, but we adapted it into a suitable test environment for our model. Our modified dataset contains Sentinel-2 L2A satellite imagery of over 4000 fields of crop of 10m spatial resolution, processed to 32x32 images. The image data from the satellite is multi-spectral (16 dimensions) and for each field, covers the growing season from early June until early August (6 time-points for each field). This type of data can be extremely valuable in helping farmers increase productivity via scalable agricultural monitoring as well as helping to reach sustainability goals \citep{cv4a}.

\begin{figure}
    \centering
    \vspace{.3in}
    \includegraphics[width=\columnwidth]{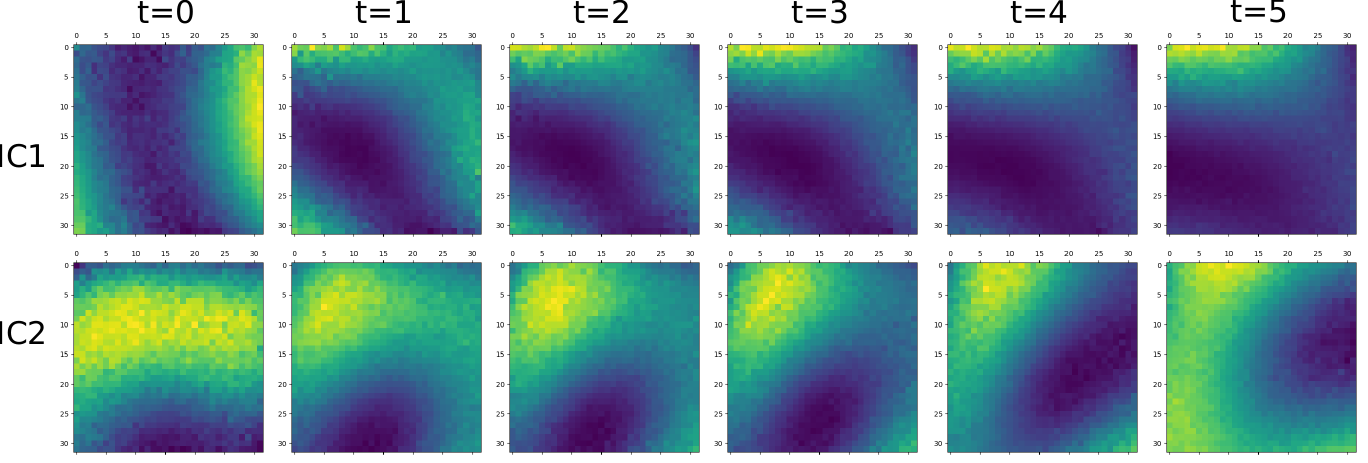}
    \caption{Examples of two nonlinear independent components inferred from the CV4A data. Horizontal axis shows how the components evolve over a field's growth-cycle.}
    \label{fig:cv4a}
\end{figure}

We first trained the tp-NICA model in a fully \textit{unsupervised} fashion on this data. To exploit both spatial and temporal dependencies over the crop growth-cycle, each independent component was taken to be spatio-temporal: 6 time-points (avg.\ 10 days apart) over the 32x32 spatial grid. 100 inducing points and the squared-exponential kernel were used. The inferred independent components were evaluated on a supervised classification task of finding the correct temporal ordering of the satellite images; that is $\sim$4000 satellite images from all of the 6 time-points were classified into those 6 time-points based only on the learned independent components using a Random Forest (RF) classifier. Figure \ref{fig:cv4a} depicts the type of spatio-temporal dependencies that the tp-NICA model learned. The results are displayed in Table~\ref{tab:super}, based on 10-fold cross-validation, to ensure generalization. The results show that tp-NICA and gp-NICA perform much better than the baseline RF classifier on the original image data. These results highlight the challenge of this task -- a successful classification requires that the learned features contain salient temporal and spatial structure that is shared across majority of the images. We also found that tp-NICA performed consistently better than gp-NICA, likely due to its ability to model any fatter tails in the data; indeed, tp-NICA models with small d.o.f.\ parameters performed the best (see Appendix). These results are very promising as remote agricultural monitoring of crops' growth cycles could have a massive benefit to farmers.

\begin{table*}
\caption{Average cross-entropies and accuracies from Random Forest (RF) classification of the images into temporal order based on the learned independent components for tp- and gp-NICA. Baseline of classifying the actual images without feature extraction (RF only) is given. l=2 and l=3 indicate the number of mixing layers assumed in the model.
}
\begin{center}
\resizebox{\textwidth}{!}{%
\begin{tabular}{@{}lllllll@{}}
\toprule
 &
  \begin{tabular}[c]{@{}l@{}}Random\\ baseline\end{tabular} &
  \begin{tabular}[c]{@{}l@{}}RF\\ only\end{tabular} &
  \begin{tabular}[c]{@{}l@{}}gp-NICA (l=2)\\ + RF\end{tabular} &
  \begin{tabular}[c]{@{}l@{}}gp-NICA (l=3)\\ + RF\end{tabular} &
  \begin{tabular}[c]{@{}l@{}}tp-NICA (l=2, nu=4)\\ + RF\end{tabular} &
  \begin{tabular}[c]{@{}l@{}}tp-NICA (l=3, nu=4)\\ + RF\end{tabular} \\ \midrule
Avg. cross-entropy &
  $1.81 \pm 0.01$ &
  $1.33 \pm 0.02$ &
  $1.19 \pm 0.01$ &
  $1.3 \pm 0.01$ &
  $\textbf{0.97} \pm 0.01$  &
  $1.08 \pm 0.02$ \\
Avg. accuracy &
  $0.17 \pm 0.01$&
  $0.27 \pm 0.01$ &
  $0.5 \pm 0.01$&
  $0.47 \pm 0.02$&
  $\textbf{0.58} \pm 0.01$&
  $0.52 \pm 0.02$ \\ \bottomrule \\
\end{tabular}
}
\label{tab:super}
\end{center}
\end{table*}


\section{RELATED WORK} \label{sec:related}
Unlike previous works, tp-NICA is built to exploit arbitrarily high-dimensional latent dependencies. It is, in particular, the first nonlinear ICA model that is built for analysing spatial and spatio-temporal data. Nevertheless, it shares several features with, and has been inspired by, several previous works.

The typical approach in existing nonlinear ICA works has been to exploit various inductive biases and latent dependencies in order to show model identifiability. One approach to this has come in the form of additional auxiliary information \citep{Hyva16NIPS, Hyva19AISTATS, Khemakhem20iVAE}. It could be argued that the tp-NICA and gp-NICA models fall under this category as the covariance kernels take as input the spatial indices, which could thus seen as auxiliary data. A critical difference is, however, that this auxiliary data is obvious and readily available. 

Another approach in nonlinear ICA has been to identify models using their latent dependencies: \citet{Halva20UAI} used a HMM of the latent components, while \citet{Hyva17AISTATS, oberhauser2021nonlinear} exploited general temporal dependencies. A model using very general dependencies was proposed by \citet{halva2021disentangling}, whose SNICA identifiability framework also subsumes our model. Despite the generality of their identifiability theory, their practical SNICA algorithm only applies to time-series or other one-dimensional dependencies. 

In terms of identifiability theory, we established the \textit{necessary} condition that GP latent components are identifiable \textit{only if} the covariance kernels of the components are distinct. This is a fundamental result as it was previously unclear whether the \textit{sufficient} conditions in \citet{halva2021disentangling} about the distinctness of the kernels were necessary or whether perhaps GPs were identifiable even more generally. 
This result follows rather directly from 
\citet{Belo97}.

On the algorithmic side, our work is most similar to \citet{ashman2020sparse, jazbec2021scalable} who define GP latent components with a neural network likelihood function. Like those authors, we employed the ideas of \citet{hensman2013gaussian, hensman2015scalable} on inducing points for computational scalability. In fact, their models are almost equivalent to the gp-NICA benchmark we used in our experiments. The tp-NICA model differs from these works in that by assuming t-process latent components, we are able to perform nonlinear ICA with components that are more generally identifiable. Also, our use of the t-process leads to a very different inference algorithm as TPs are not in the exponential family, and thus non-conjugate, hence requiring a novel approach; more details are given in the Appendix. Finally, we mention that Student-t distribution was previously considered in linear ICA by \citet{tipping2005variational}.

\paragraph{Limitations}
Because of the non-factorial posterior, the computational complexity of our full algorithm scales as $\bigO((NJ)^3)$ -- the number of independent components is typically very small, but nevertheless the multiplicative cost can be unfeasible in the rare case of very large number of components. This is not a big limitation, since this complexity can be reduced to $\bigO(J^3)$ by fully factorizing the Gaussian approximate posterior -- this is the complexity seen typically with sparse GPs, though this may come at the cost of less accurate posteriors. For our experiments this was not necessary, and non-factorized posterior provided better performance. Computational cost can be reduced also by amortizing the inference, though this again comes at the cost of an amortization gap in the lower bound. Another practical challenge is that t-Processes are non-ergodic. Considering a time signal for simplicity, a single infinitely long time-series does not have a marginal following a Student's-t distribution, since it is simply a GP re-scaled by a  single Gamma random variable. This means that multiple samples are needed for estimation and inference.

\section{CONCLUSION}
We introduced a new non-linear ICA model, tp-NICA, that uses $t$-process independent components to attain general identifiability. Unlike previous non-linear ICA algorithms, our model is designed for spatial, spatio-temporal, and any other data with high-dimensional latent structures. We also introduced a novel variational learning and inference algorithm that is able to estimate the model despite the non-conjugate structure of the t-Process prior. We also establish gp-NICA as a special case of our model but show that its identifiability is more constrained. Our results show the applicability of our approach on spatial and spatio-temporal data. We hope that our approach will inspire many further identifiable models for this type of data which is common in many important real life applications. 

\acknowledgments

H.H. received funding from the HIIT - Helsinki Institute for Information Technology. J.S. is supported by the University of Cambridge Harding Distinguished Postgraduate Scholars Programme. R.E.T. is supported by Google, Amazon, ARM, Improbable, and EPSRC grant EP/T005386/1. A.H. was funded by a Fellow Position from CIFAR and the Academy of Finland (project \#330482). The authors wish to thank the Finnish Computing Competence Infrastructure (FCCI) for supporting this project with computational and data storage resources, as well as the CSC – IT Center for Science, Finland, for generous computational resources. 

\bibliography{bib}
\bibliographystyle{apalike}

\appendix

%

%

\onecolumn
  \textit{\large Supplementary Materials for}
  
  {\large \textbf{\textit{Identifiable Feature Learning for Spatial Data with Nonlinear ICA}}

\section{Appendix: Algorithm Details}\label{apx:A}
\subsection{ELBO Derivation}
Joint pdf with pseudo-points $\ve u$:

\begin{align}
	p(\ve x, \ve s, \ve u, \ve \t) = \prod_{l=1}^m p(\ve x_l \mid \ve s_l; \ve f) \prod_{i=1}^N p(\iup{\ve s}, \iup{\ve u}\mid \iup{\t}; \iup{\thetab})p(\iup{\t}; \iup{\thetab}),
\end{align}

where for notational simplicity we have written $\ve s_l$ instead of $\ve s_{l_j}$ in the first product; also $\ve s_l = (s_l^{(1)},\dots, s_l^{(N)})$ and $\iup{\ve s} = (\iup{s_{l_1}}, \dots, \iup{s_{l_m}})$. The marginal likelihood:
\begin{align}
	\Lcal \coloneqq \log p(\ve x) = \log \int_{\ve{\t}} \int_{\ve s} \int_{\ve u}p(\ve x, \ve s, \ve u, \ve{\t}) \eqsp \rmd \ve u \rmd \ve s \rmd\ve \t,
\end{align}

is intractable but can be lower-bounded:
\begin{align}
	\Lcal &\geq \E_{q(\ve s, \ve u, \ve \t)}\left[\log \frac{p(\ve x, \ve s, \ve u \mid \ve{\t})p(\ve{\t})}{q(\ve s, \ve u, \ve \t)}  \right] \nonumber \\
	      &=\E_{q(\ve \t)}\left[\E_{q(\ve s, \ve u \mid \ve \t)}\left[\log \frac{p(\ve x, \ve s, \ve u \mid \ve{\t})}{q(\ve s, \ve u \mid \ve \t)}  \right]\right] - \kldiv{q(\ve\t)}{p(\ve{\t})} \nonumber \\
	      &=\E_{q(\ve \t)}\left[\E_{q(\ve s, \ve u \mid \ve \t)}\left[\log p(\ve x \mid\ve s, \ve u)\right]-\kldiv{q(\ve s, \ve u \mid \ve \t)}{p(\ve s, \ve u \mid \ve \t)}\right] - \kldiv{q(\ve\t)}{p(\ve{\t})} \nonumber \\
	      &\text{choosing to approximate } q(\ve s \mid \ve u, \ve \t)\coloneqq p(\ve s \mid \ve u, \ve \t)\,\,\text{results in a further lower bound:}\nonumber\\
	      &\geq \E_{q(\ve \t)}\left[\E_{q(\ve u \mid \ve \t)}\left[\E_{p(\ve s \mid \ve u, \ve \t)}\left[\log p(\ve x \mid \ve s)\right]\right]-\kldiv{q(\ve u \mid \ve \t)}{p(\ve u \mid \ve \t)}\right] - \kldiv{q(\ve\t)}{p(\ve{\t})} \nonumber \\
	      &= \E_{q(\ve \t)}\left[\underbrace{\E_{\tilde{q}(\ve s \mid \ve \t)}\left[\log p(\ve x \mid \ve s)\right]}_{(\ast)}-\underbrace{\kldiv{q(\ve u \mid \ve \t)}{p(\ve u \mid \ve \t)}}_{(\ast\ast)}\right] - \underbrace{\kldiv{q(\ve\t)}{p(\ve{\t})}}_{(\ast\ast\ast)} \label{eq:elbo_fin},
\end{align}
with 
\begin{align}
\tilde{q}(\ve s\mid \ve \t) \coloneqq \int p(\ve s \mid \ve u, \ve \t)q(\ve u \mid \ve \t) \rmd \ve u\end{align}

\subsection{Distribution of Latent Components} \label{apx:distrib_latents}
Note that the pseudo-points are like additional latent variables at some extra locations, and therefore we have:
\begin{align}
p(\ve s, \ve u \mid \ve \t) &= \prod_{i=1}^N p(\iup{\ve s}, \iup{\ve u}\mid \iup{\t}; \iup{\thetab}) \nonumber \\
			    &= \prod_{i=1}^N (2\pi)^{-\frac{m+J}{2}} \lvert  \iup{\m K}_{\ve r} \rvert^{-\frac{1}{2}}\exp\left\{-\frac{1}{2}(\iup{\ve r})\tran \m K_{\ve r}^{(i)-1} \iup{\ve r} \right\},
\end{align}
where
\begin{align*}
	\iup{\ve r} &= \begin{pmatrix} \iup{\ve s} \\\iup{\ve u}\end{pmatrix}\eqsp\eqsp, \iup{\m K}_{\ve r} = \begin{pmatrix} \iup{\m K}_{\ve s, \ve s'} & \iup{\m K}_{\ve s, \ve u'} \\ \iup{\m K}_{\ve u, \ve s'} & \iup{\m K}_{\ve u, \ve u'}\end{pmatrix},
\end{align*}
and, for brevity, we define $\iup{\m K}\coloneqq \frac{\m K_{\iup{\thetab}}}{\iup{\t}}$. We will re-order elements for later use; that is we define $\ve r \coloneqq (\ve s_1, \dots, \ve s_m, \ve u_1, \dots, \ve u_J)=(\ve s, \ve u)$, which results in a covariance matrix:
\begin{align}\label{eq:Kr}
	\Sigmab = \begin{pmatrix}
		\m K_{\ve s, \ve s'} & \m K_{\ve s, \ve u'} 	    \\
		\m K_{\ve u, \ve s'} & \m K_{\ve u, \ve u'}
	\end{pmatrix},
\end{align}

where $\m K_{\ve s, \ve s'}$ is $mN\times mN$, $\m K_{\ve u, \ve u'}$ is $JN \times JN$, and  $\m K_{\ve s, \ve u'}$ is $mN \times JN$. These block matrices also have a specific form; using $\m K_{\ve s, \ve s'}$ as an example, this block can be considered as $m \times m$ matrix of $N \times N$ blocks as its elements where each of the blocks is a diagonal matrix since the latent variables factorize over $N$. In particular, the form is (dropping subscripts for convenience):
\begin{align}\label{eq:K}
\m K = \begin{pmatrix}
		\m K_{1,1} &\cdots & \m K_{1,m} 	    \\
		\vdots   & \ddots & \\
			\m K_{m,1}   & & \m K_{m,m}
		\end{pmatrix},
\end{align}
where $\m K_{i,j} = \diag(k_{i,j}^{(1)},\dots, k_{i,j}^{(N)})$. We can then write the joint as a big block-diagonal Gaussian:
\begin{align}
	p(\ve s, \ve u \mid \ve \t) &= (2\pi)^{-\frac{N(m+J)}{2}} \lvert \Sigmab \rvert^{-\frac{1}{2}}\exp\left\{-\frac{1}{2}\ve r\tran \Sigmab^{-1} \ve r \right\}.
\end{align}
Using the properties of partitioned Gaussians gives the conditional distribution as:
\begin{align}\label{eq:condit}
	p(\ve s \mid \ve u, \ve \t) \propto \exp\left\{(\ve s-\ve \mu_{\ve s \mid \ve u})\tran \Sigmab_{\ve s \mid \ve u}^{-1}(\ve s-\ve \mu_{\ve s \mid \ve u})\right\}, 
\end{align}
where
\begin{align}
	\ve \mu_{\ve s \mid \ve u} &= \m K_{\ve s, \ve u'} \m K_{\ve u, \ve u'}^{-1}\ve u \\
	\Sigmab_{\ve s \mid \ve u} &= \m K_{\ve s, \ve s'} - \m K_{\ve s, \ve u'}\m K_{\ve u, \ve u'}^{-1} \m K_{\ve s, \ve u'}\tran,
\end{align}
And the marginal distribution:
\begin{align} \label{eq:marginal}
	p(\ve u \mid \ve \t) \propto \exp\left\{\ve u\tran \m K_{\ve u, \ve u'}^{-1}\ve u\right\}.
\end{align}

\subsection{Approximate Posterior}
We assume following form of the approximate posterior:
\begin{align}
	q(\ve u, \ve \t) &= q(\ve u\mid \ve \t; \phib, \thetab)q(\ve \t; \lambdab) \nonumber \\ 
	&\coloneqq \frac{\prod_{j=1}^J \psi(\ve u_j; \phib_j)\prod_{i=1}^N p(\iup{\ve u} \mid \iup{\t}; \iup{\thetab})}{Z(\phib, \thetab, \ve \t)}\prod_{i=1}^N q(\iup{\t}; \iup{\lambda}),
\end{align}
where $J$ is the number of pseudo-locations, $\ve u_j = (\iup{u_j})_{i=1:N}$, and $\iup{\ve u}= (\iup{u}_j)_{j=1:J}$. The approximate likelihood is assumed to be a Gaussian factor: 
\begin{align}\label{eq:approx_likelih}
	\prod_{j=1}^J \psi(\ve u_j; \phib_j) &= \prod_{j=1}^J \exp\left\{-\frac{1}{2}(\ve u_j\tran \tilde{\ve W}_j\tran \tilde{\ve W}_j\ve u_j -2\ve u_j\tran \tilde{\ve W}_j\tran\tilde{\ve y}_j)\right\}.
\end{align}
Now define:
\begin{align}
	\Lambdab_j &= \tilde{\ve W}_j\tran \tilde{\ve W}_j,\nonumber \\
	\ve m_j &= \tilde{\ve W}_j\tran\tilde{\ve y}_j,
\end{align}
Next, recall $\ve u = (\ve u_1,\dots, \ve u_J)$, and define a block-diagonal matrix $\Lambdab \coloneqq \diag(\Lambdab_1,\dots, \Lambdab_J)$, which allows us to simplify above into:
\begin{align}\label{eq:lh_term}
	\prod_{j=1}^J \psi(\ve u_j; \phib_j) &= \exp\left\{-\frac{1}{2}(\ve u\tran \Lambdab \ve u -2\ve u\tran \ve m)\right\}.
\end{align}
With \eqref{eq:lh_term} and \eqref{eq:marginal}, we thus have:
\begin{align*}
	q(\ve u\mid \ve \t;\phib, \thetab) &\propto \exp\left\{-\frac{1}{2}(\ve u\tran \Lambdab \ve u -2\ve u\tran \ve m)\right\} (2\pi)^{-\frac{NJ}{2}}\lvert \m K_{\ve u, \ve u'} \rvert^{-\frac{1}{2}}\exp\left\{-\frac{1}{2} \ve u \tran \m K_{\ve u, \ve u'}^{-1} \ve u \right\} \\
							    &= (2\pi)^{-\frac{NJ}{2}}\lvert \m K_{\ve u, \ve u'} \rvert^{-\frac{1}{2}}\exp\left\{-\frac{1}{2}(\ve u\tran (\Lambdab+\m K_{\ve u, \ve u'}^{-1}) \ve u -2\ve u\tran \ve m)\right\}.
\end{align*}
Simplify with $\m J = (\m K_{\ve u, \ve u'}^{-1}+\Lambdab)$ and $\ve h = \m J^{-1}\ve m $:
\begin{align}
	q(\ve u\mid \ve \t;\phib, \thetab) &\propto (2\pi)^{-\frac{NJ}{2}}\lvert \m K_{\ve u, \ve u'} \rvert^{-\frac{1}{2}}\exp\left\{-\frac{1}{2}(\ve u\tran \m J \ve u -2\ve u\tran \m J \ve h)\right\}.
\end{align}
The normalizer can be calculated as:
\begin{align} \label{eq:normalizer}
	Z(\phib, \thetab, \ve \t) &= (2\pi)^{-\frac{NJ}{2}}\lvert \m K_{\ve u, \ve u'} \rvert^{-\frac{1}{2}}\int \exp\left\{-\frac{1}{2}(\ve u\tran \m J \ve u -2\ve u\tran \m J \ve h)\right\} d\ve u \nonumber\\
						   &=(2\pi)^{-\frac{NJ}{2}}\lvert \m K_{\ve u, \ve u'} \rvert^{-\frac{1}{2}}(2\pi)^{\frac{NJ}{2}}\lvert \m J^{-1} \rvert^{\frac{1}{2}}\exp\{\frac{1}{2}\ve h\tran \m J \ve h \} \nonumber\\
						   &=\lvert \m K_{\ve u, \ve u'} \rvert^{-\frac{1}{2}}\lvert \m J \rvert^{-\frac{1}{2}}\exp\{\frac{1}{2}\ve m\tran \m J^{-1} \ve m \} \nonumber\\
						   &=\lvert \Lambdab\m K_{\ve u, \ve u'} + \m I \rvert^{-\frac{1}{2}} \exp\{\frac{1}{2}\ve m\tran \m J^{-1} \ve m \} \nonumber \\
	\Rightarrow \log Z &= \frac{1}{2}\ve m\tran \m K_{\ve u, \ve u'}(\Lambdab \m K_{\ve u, \ve u'} + \m I)^{-1} \ve m  -\frac{1}{2}\log\lvert \Lambdab\m K_{\ve u, \ve u'} + \m I \rvert
\end{align}
And for completeness:
\begin{align} \label{eq:qs}
 q(\ve u\mid \ve \t;\phib, \thetab) =\lvert 2\pi \m J^{-1} \rvert^{-\frac{1}{2}}\exp\left\{-\frac{1}{2}(\ve u - \ve h)\tran \m J (\ve u - \ve h) \right\}.
\end{align}

\subsection{Computing \texorpdfstring{$(\ast)$}{(*)}} \label{apx:compute_ast}
We have: 
\begin{align} \label{eq:qut}
\tilde{q}(\ve s\mid \ve \t) \coloneqq \int p(\ve s \mid \ve u, \ve \t)q(\ve u \mid \ve \t) \rmd \ve u.
\end{align}
Substitute in \eqref{eq:condit} and \eqref{eq:qs}:
\begin{align}
	\tilde{q}(\ve s\mid \ve \t) \propto \int \exp\left\{(\ve s-\ve \mu_{\ve s \mid \ve u})\tran \Sigmab_{\ve s \mid \ve u}^{-1}(\ve s-\ve \mu_{\ve s \mid \ve u})\right\} \exp\left\{-\frac{1}{2}(\ve u - \ve h)\tran \m J (\ve u - \ve h) \right\} \rmd \ve u,
\end{align}
so that the \textit{approximate} marginal is a Gaussian:
\begin{align}
	\tilde{q}(\ve s\mid \ve \t) &\propto \exp\left\{(\ve s - \tilde{\ve \mu})\tran \tilde{\Sigmab}^{-1} (\ve s - \tilde{\ve \mu}) \right\},
\end{align}
with:
\begin{align}
	\tilde{\ve \mu} &= \m K_{\ve s, \ve u'}\m K_{\ve u, \ve u'}^{-1}\ve h  \nonumber\\
			&= \m K_{\ve s, \ve u'}\m K_{\ve u, \ve u'}^{-1}(\m K_{\ve u, \ve u'}^{-1}+\Lambdab)^{-1} \ve m \nonumber \\
			&= \m K_{\ve s, \ve u'}(\m K_{\ve u, \ve u'}+\Lambdab^{-1})^{-1}\Lambdab^{-1} \ve m \nonumber \\
			&= \m K_{\ve s, \ve u'}(\m I +\Lambdab\m K_{\ve u, \ve u'})^{-1} \ve m \\
	\tilde{\Sigmab} &= \m K_{\ve s, \ve s'} - \m K_{\ve s, \ve u'}\m K_{\ve u, \ve u'}^{-1} \m K_{\ve u, \ve s} + \m K_{\ve s, \ve u'}\m K_{\ve u, \ve u'}^{-1} \m J^{-1} \m K_{\ve u, \ve u'}^{-1}\m K_{\ve u, \ve s'} \nonumber \\
			&= \m K_{\ve s, \ve s'} - \m K_{\ve s, \ve u'}(\m K_{\ve u, \ve u'}^{-1}  - \m K_{\ve u, \ve u'}^{-1} \m J^{-1} \m K_{\ve u, \ve u'}^{-1})\m K_{\ve u, \ve s'} \nonumber \\
			&= \m K_{\ve s, \ve s'} - \m K_{\ve s, \ve u'}(\m K_{\ve u, \ve u'}^{-1}  - \m K_{\ve u, \ve u'}^{-1} (\Lambdab+\m K_{\ve u, \ve u'}^{-1})^{-1} \m K_{\ve u, \ve u'}^{-1})\m K_{\ve u, \ve s'} \nonumber  \\
			&= \m K_{\ve s, \ve s'} - \m K_{\ve s, \ve u'}(\m K_{\ve u, \ve u'} + \Lambdab^{-1})^{-1} \m K_{\ve u, \ve s'} \nonumber  \\
			&= \m K_{\ve s, \ve s'} - \m K_{\ve s, \ve u'}(\m I+ \Lambdab \m K_{\ve u, \ve u'})^{-1}\Lambdab  \m K_{\ve u, \ve s'}
\end{align}
We can sample from this distribution to approximate $(\ast)$ and take gradients using the reparametrization trick. Further, note that in practice we only need the $m$, $N\times N$, diagonal blocks of the above covariance matrix as we only have to sample from $\tilde{q}(\ve s_l\mid \ve \t)\,\,\,\, \forall t$, rather than the full joint.
\subsection{Computing \texorpdfstring{$(\ast\ast)$}{(**)}}\label{apx:compute_astast}
\begin{align}
	\kldiv{q(\ve u \mid \ve \t)}{p(\ve u \mid \ve \t)} &= \int_{\ve u}q(\ve u \mid \ve \t)\log\frac{q(\ve u \mid \ve \t)}{p(\ve u \mid \ve \t)} \\
	&=\int_{\ve u}q(\ve u \mid \ve \t)\log\frac{\prod_{j=1}^J \psi(\ve u_j; \phib_j)\prod_{i=1}^N p(\iup{\ve u} \mid \iup{\t}; \iup{\thetab})}{Z(\phib, \thetab, \ve \t)\prod_{i=1}^N p(\iup{\ve u} \mid \iup{\t}; \iup{\thetab})} \\
	&=\sum_{j=1}^J\E_{q(\ve u_j \mid \ve \t)}\left[\log \psi(\ve u_j; \phib_j)\right] - \log Z(\phib,\thetab, \ve \t),
\end{align}

where $\log Z(\phib, \ve \mu, \thetab, \ve \t)$ can be computed using \eqref{eq:normalizer}, while:
\begin{align}
	&\sum_{j=1}^J\E_{q(\ve u_j \mid \ve \t)}\left[\log \psi(\ve u_j; \phib_j)\right] =\sum_{j=1}^J \E_{q(\ve u_j \mid \ve \t)}\left[-\frac{1}{2}\ve u_j\tran \Lambdab_{j} \ve u_j+\ve u_j\tran \ve m_j\right] \nonumber\\
											&=\sum_{j=1}^J -\frac{1}{2}\Tr(\Lambdab_{j}[\m J^{-1}]_j) -\frac{1}{2}\ve h_j\tran \Lambdab_j \ve h_j +\ve h_j\tran \ve m_j \nonumber \\
											&=\sum_{j=1}^J -\frac{1}{2}\Tr(\Lambdab_{j}[\m K_{\ve u, \ve u}(\Lambdab\m K_{\ve u, \ve u} + \m I)^{-1}]_j) -\frac{1}{2}\ve h_j\tran \Lambdab_j \ve h_j +\ve h_j\tran \ve m_j \nonumber \\
											&=\sum_{j=1}^J -\frac{1}{2}\Tr([\m K_{\ve u, \ve u}(\Lambdab\m K_{\ve u, \ve u} + \m I)^{-1}\Lambdab]_j) -\frac{1}{2}\ve h_j\tran \Lambdab_j \ve h_j +\ve h_j\tran \ve m_j \nonumber 
\end{align}

\subsection{Computing \texorpdfstring{$(\ast \ast \ast)$}{(***)}}
We will make $\kldiv{q(\ve\t)}{p(\ve{\t})}$ tractable by assuming $q(\ve \t)$ to follow factorial gamma distribution and using properties of exponential families:
\begin{align}
	\kldiv{q(\ve\t)}{p(\ve{\t})} &= \sum_{i=1}^N \kldiv{q(\iup{\t}; \iup{\lambda})}{p(\iup{\t}; \iup{\gamma})} \\
				  &=\sum_{i=1}^N A(\iup{\ve \eta}_p)-A(\iup{\ve \eta}_q) - (\iup{\ve \eta}_p - \iup{\ve \eta}_q)\tran \nabla A(\iup{\ve \eta}_q),
\end{align}
where $\iup{\ve \eta}$ are the natural parameters, and $A(\cdot)$ is the log-normalizer and gradient of the log normalizer gives the mean parameters.

\clearpage

\section{Appendix: Proofs of Theorems}\label{apx:B}
\paragraph{Proof of Theorem \ref{theo:main}} (A1) requires the noise-free data $\ve f(\ve s)$ to have tails that are only slightly heavier than Gaussian, while (A2) and (A3) ensure non-degeneracy and non-Gaussianity of the noise-free data, respectively. With these assumptions, Theorem 1 of \citet{halva2021disentangling} shows that one can identify the noise-free mixture from the noisy mixture: $p(\ve x) = \widetilde{p}(\ve x) \implies p(\ve f(\ve s))=p(\widetilde{\ve f}(\ve s))$. For our tp-NICA model, the noise-free distribution can be written as (assuming zero-mean TP):
\begin{equation}
\prod_{j=1}^m |\m J \ve g(\ve x_l)| \prod_{i=1}^N \pi^{-\frac{d}{2}}\lvert \m \Sigma^{(i)} \rvert^{-\frac{1}{2}} \nu^{{(i)} \frac{\nu^{(i)}}{2}} \frac{\Gamma(\frac{\nu^{(i)}+d}{2})}{\Gamma(\frac{\nu^{(i)}}{2})}\left(\nu^{(i)} + \ve 
    g^{(i)}(\ve x_{1:m})^T \m \Sigma^{-1,{(i)}} \ve g^{(i)}(\ve x_{1:m})\right)^{-\frac{\nu^{(i)}+d}{2}},
\end{equation}
where $\m J$ is the Jacobian resulting from the change of variable, and $\ve g=(\ve g^{(1)}, \dots, \ve g^{(N)})=\ve f^{-1}$. In other words, $\ve g^{(i)}(\ve x_{1:m}) = \ve s^{(i)} =(s_1^{(i)},\dots, s_m^{(i)})$.

Define then $Q^{(i)}:=\log p^{(i)}$, so that
\begin{align} 
\frac{\partial}{\partial s_{l_k}^{(i)}} Q^{(i)} &={-\frac{\nu^{(i)}+d}{(\nu^{(i)} + \ve s^{{(i)}^T} \m \Sigma^{-1, (i)} \ve s^{(i)})}} \ve s^{(i)T} \m \Sigma_{[:, k]}^{-1, (i)} \\
\frac{\partial^2}{\partial s_{l_k}^{(i)}\partial s_{l_j}^{(i)}} Q^{(i)}& ={-\frac{\nu^{(i)}+d}{(\nu^{(i)} + \ve s^{{(i)}^T} \m \Sigma^{-1, (i)} \ve s^{(i)})}} \Sigma_{[j, k]}^{-1, (i)} \nonumber \\ 
&{+2\frac{\nu^{(i)}+d}{(\nu^{(i)} + \ve s^{{(i)}^T} \m \Sigma^{-1, (i)} \ve s^{(i)})^2}}\ve s^{(i)T} \m \Sigma_{[:, k]}^{-1, (i)} \ve s^{(i)T} \m \Sigma_{[:, l]}^{-1, (i)}.
\end{align}
The $s_{l_k}^{(i)}, s_{l_j}^{(i)}$ don't factorize in above, and thus we can not rewrite this as:
\begin{equation}
    \frac{\partial^2}{\partial s_{l_k}^{(i)} \partial s_{l_j}^{(i)}} Q^{(i)}
	= c \, \alpha (s_{l_k}^{(i)}, \ve s_{(-l_k,-l_j)}^{(i)}) \alpha (s_{l_j}^{(i)}, \ve s_{(-l_k,-l_j)}^{(i)})\eqsp
\end{equation}
In other words, the distribution is not \textit{quasi}-Gaussian. This is precisely the condition needed to satisfy assumptions of Theorem 2 in \citet{halva2021disentangling} and by this theorem $p(\ve f(\ve s))=p(\widetilde{\ve f}(\ve s)) \implies \ve f^{-1} \sim \widetilde{\ve f}^{-1}$, where $\sim$ denotes equivalence up to permutation and coordinate-wise bijective transformation of the elements of the de-mixing function $\ve f^{-1}$.


\paragraph{Proof of Theorem \ref{theo:nec}}
Theorem 5 in \citet{halva2021disentangling} shows that if the covariance kernel functions of the components are distinct, the components are identifiable (hence sufficiency is already satisfied). It remains to prove necessity. To do this, it is enough to show that the condition is necessary for any specific $\ve f$ -- indeed it is enough to show the necessity for linear ICA, which we do by adapting its well-known theory \citep{Belo97}. We proceed by contradiction: Assume there exist an identifiable model with a linear mixing function and components with non-distinct kernels $\kappa^{(i)} = \kappa^{(j)}$, that is, some two components $i$  and $j$ have the same covariance kernel. Next, consider the alternative components obtained by an orthogonal rotation: $s'^{(i)}=\frac{1}{\sqrt{2}}[s^{(i)}+ s^{(j)}]$ and $s'^{(j)}=\frac{1}{\sqrt{2}}[s^{(i)}- s^{(j)}]$. These new components have the same autocorrelations as the original components. They are also uncorrelated, and hence independent. The Jacobian of the orthogonal transformation is identity so, the likelihood of these components is equal to the likelihood of the original components. This implies unidentifiability of the components. Distinct kernels are therefore necessary in this specific example. Thus, if distinctness of kernels functions is violated, nonlinear mixing cannot be identifiable in general.

\section{Appendix: Experiment Details}

\paragraph{Experiments on Simulated Data}
In total 1024 independent samples were used to train each model specification, that is the training data had 1024 samples of $M \times 32 \times 32$ observations where each was generated by a nonlinear mixing matrix of $N \times 32 \times 32$ independent components, where $M$ and $N$ are the data dimension and the number of independent components. Output noise was added to the observations such that it accounted to approximately $10\%$ of the latent components overall variance. The training of these samples was done in minibatches of 8 samples. This entire procedure was repeated 6 times from different random seeds to ensure reproducibility, the variance between the seeds' results is depicted by error bars in the Figures \ref{fig:sim}. In these simulations the d.o.f. of the tp-NICA was set to four to ensure fat tails and thus ensure that the data generated is different from gp-NICA. Estimation of the d.o.f. parameter, $\nu$, is in general known to be potentially ill-defined \citep{FernandezPitfalls} and thus we set $\nu$ to the ground-truth values in the simulation as it was not of major focus in our paper (though in our experience, estimating it resulted nearly identical results). All other parameters were fully estimated. Adam optimizer was used with learning rates of $1e^{-1}$ and $1e^{-4}$ for the variational parameters and model parameters respectively. In order to make our code as stable and computationally efficient as possible, matrix inverses were never performed directly, but rather always using linear solves of matrix-vector products. The most important factor in computational efficacy of the model is the setting on number of pseudo-points used. We found that 50-points provided appropriate balance between speed and good identifiability results. Typical training time of our model was around 12-24h on either a single A100 or V100 NVIDIA GPU. All further details can be found in our code at [redacted for anonymity]. 

\begin{figure}
    \centering
    \vspace{.3in}
    \includegraphics[width=0.4\columnwidth]{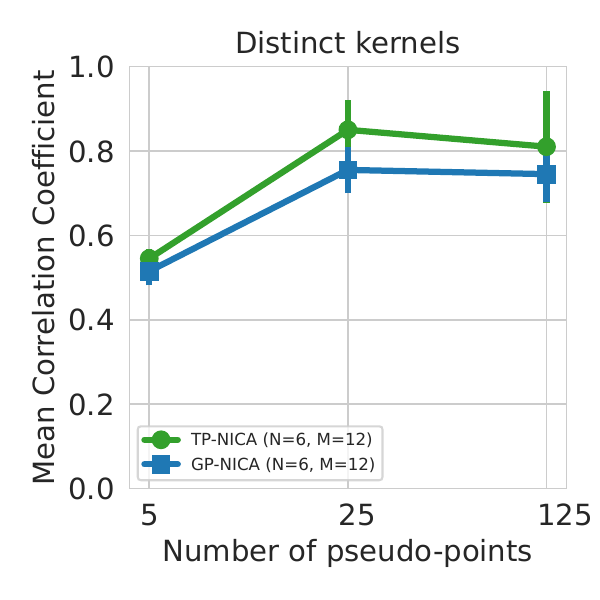}
    \vspace{.3in}
    \caption{Simulation results for different number of pseudo-points for tp-NICA and gp-NICA models with three mixing layers and assuming distinct kernels.}
    \label{fig:sim_pseudo}
\end{figure}

}

\paragraph{Experiments on CV4A Satellite Data}
The Computer Vision for Agriculture (CV4A) Kenya Crop Type dataset \citep{cv4a} was originally designed for an supervised classification task of different field crops and was published at the ICLR 2020. We adapted this data to our use by considering the data in unsupervised sense after first cropping the satellite images into 32x32 images (here were benefited from the data processing code available at a public github  \href{https://github.com/radiantearth/crop-type-detection-ICLR-2020/blob/master/solutions/KarimAmer/prepare_data.py}{repository} that created the crops and normalized the images by their mean values and standard deviation. The 16 dimensional satellite image data from the Sentinel-12 satellite consists of the following bands: ultra blue, blue, red, green, five visible and near infrared wavelengths, four short-wave infrared wavelengths, cloud probability layer, and vegetation index layers constructed from the other layers.

The \href{www.com}{redacted} repository for our paper also contains all the further details for our experiments on this data. 

\section{Appendix: More Discussion on Related Work} \label{related.supp}

We explain in more detail the difference to \citet{ashman2020sparse, jazbec2021scalable}. TPs are not in the exponential family, and thus non-conjugate; we solved this  by re-phrasing our objective in terms of infinite mixture of GPs scaled by a Gamma random variable. Furthermore, unlike those two papers, we used free-form variational parameters \citep{ranganath14} rather than amortizing our inference; we utilize the reparametrization trick in optimizing these free-form parameters, thus combining the strengths of BBVI \citep{ranganath14} and VAE\citep{kingma2014autoencoding} approaches. Finally, unlike \citet{ashman2020sparse, jazbec2021scalable} and most other previous works in VAEs, we do not assume a factored posterior approximation, but rather assume that the approximate likelihood term only factors over time but not between components -- this allows us to parameterize it in terms of block diagonal matrix of Cholesky factors, where blocks size equals the number of components.

\vfill

\end{document}